\documentclass[11pt]{article}
\usepackage{acl2014}
\usepackage{times}
\usepackage{url}
\usepackage{latexsym}
\usepackage{amsmath}
\usepackage{multirow}
\usepackage{url}
\usepackage{array}
\usepackage{graphicx}
\usepackage[T5,T1]{fontenc}
\usepackage{algorithmic, amssymb, amsmath}   
\usepackage{algorithm}
\usepackage{multirow}
\usepackage{comment}

\title{Automatically constructing Wordnet  synsets}

\author{Khang Nhut Lam, Feras Al Tarouti and Jugal Kalita \\
Computer Science department \\
University of Colorado \\
1420 Austin Bluffs Pkwy, Colorado Springs, CO 80918, USA \\
{\tt \{klam2,faltarou,jkalita\}@uccs.edu}\\}

\date{}

\begin{document}

\maketitle
\begin{abstract}
Manually constructing a Wordnet is a difficult task, needing years of experts' time. As a first step to automatically construct full Wordnets, we propose approaches to generate Wordnet synsets for languages both resource-rich and resource-poor, using publicly available Wordnets, a machine translator and/or a single bilingual dictionary. Our algorithms translate synsets of existing Wordnets to a target language \textit{T}, then apply a ranking method on the translation candidates to find best translations in \textit{T}. Our approaches are applicable to any language which has at least one existing bilingual dictionary translating from English to it.
\end{abstract}

\section{Introduction}

Wordnets are intricate and substantive repositories of lexical knowledge and have become important resources for computational processing of natural languages and for information retrieval. Good quality Wordnets are available only for a few "resource-rich" languages such as English and Japanese. Published approaches to automatically build new Wordnets are manual or semi-automatic and can be used only for languages that already possess some lexical resources. 

The Princeton Wordnet (PWN) \cite{Fellbaum1998} was painstakingly constructed manually over many decades. Wordnets, except the PWN, have been usually constructed by one of two approaches. The first approach translates the PWN to \emph{T} \cite{Bilgin2004}, \cite{Barbu2005}, \cite{Kaji2006}, \cite{Sagot2008}, \cite{Saveski2010} and \cite{Oliver2012}; while the second approach builds a Wordnet in \emph{T}, and then aligns it with the PWN by generating translations \cite{Gunawan2010}. In terms of popularity, the first approach dominates over the second approach. Wordnets generated using the second approach have different structures from the PWN; however, the complex agglutinative morphology, culture specific meanings and usages of words and phrases of target languages can be maintained. In contrast, Wordnets created using the first approach have the same structure as the PWN. 

One of our goals is to automatically generate high quality synsets, each of which is a set of cognitive synonyms, for Wordnets having the same structure as the PWN in several languages. Therefore, we use the first approach to construct Wordnets. This paper discusses the first step of a project to automatically build core Wordnets for languages with low amounts of resources (viz., Arabic and Vietnamese), resource-poor languages (viz., Assamese) or endangered languages (viz., Dimasa and Karbi)\footnote{ISO 693-3 codes of Arabic, Assamese, Dimasa, Karbi and Vietnamese are \emph{arb, asm, dis, ajz} and \emph{vie}, respectively.}. The sizes and the qualities of freely existing resources, if any, for these languages vary, but are not usually high. Hence, our second goal is to use a limited number of freely available resources in the target languages as input to our algorithms to ensure that our methods can be felicitously used with languages that lack much resource. In addition, our approaches need to have a capability to reduce noise coming from the existing resources that we use. For translation, we use a free machine translator (MT) and restrict ourselves to using it as the only "dictionary" we can have. For research purposes, we have obtained free access to the Microsoft Translator, which supports translations among 44 languages. In particular, given public Wordnets aligned to the PWN ( such as  the FinnWordNet (FWN) \cite{Linden2010} and the JapaneseWordNet (JWN) \cite{Hitoshi2008} ) and the Microsoft Translator, we build Wordnet synsets for \emph{arb, asm, dis, ajz} and \emph{vie}.  

\section{Proposed approaches}

In this section, we propose approaches to create Wordnet synsets for a target languages \emph{T} using existing Wordnets and the MT and/or a single bilingual dictionary. We take advantage of the fact that every synset in PWN has a unique \emph{offset-POS}, referring to the offset for a synset with a particular  part-of-speech (POS) from the beginning of its data file. Each synset may have one or more words, each of which may be in one or more synsets. Words in a synset have the same sense. The basic idea is to extract corresponding synsets for each \emph{offset-POS} from existing Wordnets linked to PWN, in several languages. Next, we translate extracted synsets in each language to \emph{T} to produce so-called \emph{synset candidates} using MT. Then, we apply a ranking method on these candidates to find the correct words for a specific \emph{offset-POS} in \emph{T}.  

\subsection{Generating synset candidates}
We propose three approaches to generate synset candidates for each \emph{offset-POS} in \emph{T}.

\subsubsection{The direct translation (DR) approach}
\begin{figure}[!h]
\centering \includegraphics[width=0.48\textwidth]{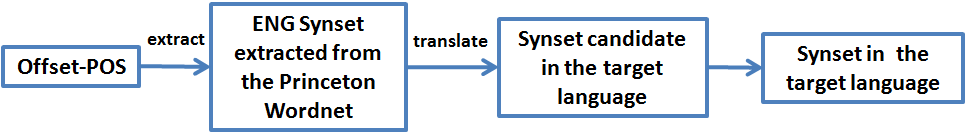}\vspace{-0.05 in}
\caption{The DR approach to construct Wordnet synsets in a target language \emph{T}.}
\label{fig:Algorithm1}
\end{figure}

The first approach directly translates synsets in PWN to \emph{T} as in Figure \ref{fig:Algorithm1}. 

For each \emph{offset-POS}, we extract words in that synset from the PWN and translate them to the target language to generate translation candidates.  

\subsubsection{Approach using intermediate Wordnets (IW)}
\begin{figure}[!h]
\centering \includegraphics[width=0.5\textwidth]{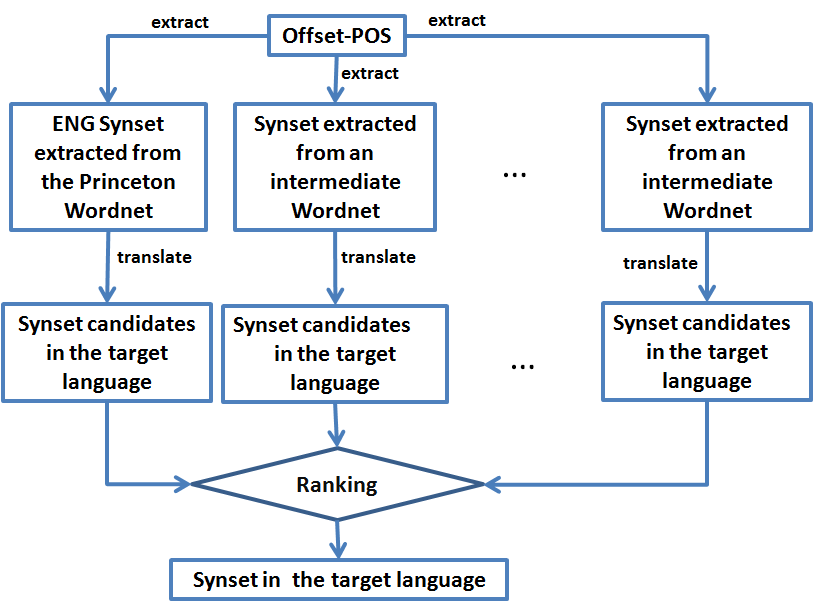}\vspace{-0.05 in}
\caption{ The IW approach to construct Wordnet synsets in a target language \emph{T}}
\label{fig:Algorithm2}
\end{figure}

To handle ambiguities in synset translation, we propose the IW approach as in Figure \ref{fig:Algorithm2}. Publicly available Wordnets in various languages, which we call intermediate Wordnets, are used as resources to create synsets for Wordnets. For each \emph{offset-POS}, we extract its corresponding synsets from intermediate Wordnets. Then, the extracted synsets, which are in different languages, are translated to \emph{T} using MT to generate synset candidates. Depending on which Wordnets are used and the number of intermediate Wordnets, the number of candidates in each synset and the number of synsets in the new Wordnets change.

\vspace{-4mm}
\vspace{-4mm}

 \subsubsection{Approach using intermediate Wordnets and a dictionary (IWND)}
\begin{figure}[!h]
\centering \includegraphics[width=0.50\textwidth]{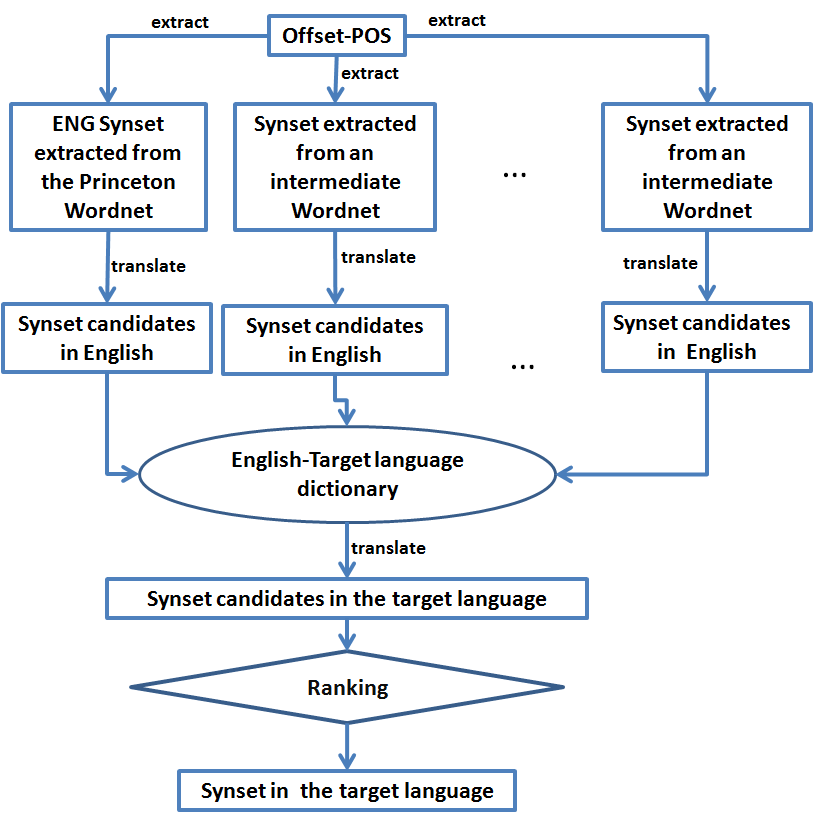}
\caption{The IWND approach to construct Wordnet synsets}
\label{fig:Algorithm3}
\end{figure}

The IW approach for creating Wordnet synsets decreases ambiguities in translations. However, we need more than one bilingual dictionary from each intermediate languages to \emph{T}. Such dictionaries are not always available for many languages, especially the ones that are resource poor. The IWND approach is like the IW approach, but instead of translating immediately from the intermediate languages to the target language, we translate synsets extracted from intermediate Wordnets to English (\emph{eng}), then translate them to the target language. The IWND approach is presented in Figure \ref{fig:Algorithm3}.
 
\subsection{Ranking method}
For each of \emph{offset-POS}, we have many translation candidates. A translation candidate with a higher rank is more likely to become a word belonging to the corresponding \emph{offset-POS} of the new Wordnet in the target language. Candidates having the same ranks are treated similarly. The rank value in the range 0.00 to 1.00. The rank of a word \emph{w}, the so-called $rank_w$, is computed as below. 

\vspace{0.8mm}
$rank_w= \frac{occur_w}{numCandidates} * \frac{numDstWordnets}{numWordnets}$  where:
\vspace{-0.5mm}
\begin{itemize} 
\setlength{\itemsep}{1pt}
\setlength{\parskip}{0pt}
\setlength{\parsep}{0pt}
\renewcommand\labelitemi{-}
\item \emph{numCandidates} is the total number of translation candidates of an \emph{offset-POS}
\item $occur_w$ is the occurrence count of the word \emph{w} in the \emph{numCandidates}
\item  \emph{numWordnets} is the number of intermediate Wordnets used, and
\item \emph{numDstWordnets} is the number of distinct intermediate Wordnets that have words translated to the word \emph{w} in the target language.
\end{itemize}
\vspace{-0.05 in}

Our motivation for this rank formula is the following. If a candidate has a higher occurrence count, it has a greater chance to become a correct translation. Therefore, the occurrence count of each candidate needs to be taken into account. We normalize the occurrence count of a word by dividing it by \emph{numCandidates}. In addition, if a candidate is translated from different words having the same sense in different languages, this candidate is more likely to be a correct translation. Hence, we multiply the first fraction by \emph{numDstWordnets}. To normalize, we divide results by the number of intermediate Wordnet used.

\begin{figure}[!h]
\centering \includegraphics[width=0.50\textwidth]{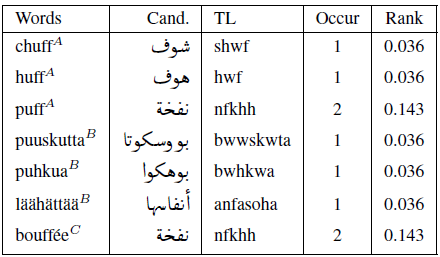}\vspace{-0.05 in}
\caption{Example of calculating the ranks of candidates translated from words belonging to the \emph{offset-POS} "00006802-v" in 4 Wordnets: PWN, FWN, JWN and WWN. The $word^A$, $word^B$ and $word^C$ are obtained from PWN, FWN and WWN, respectively. The JWN does not contain this \emph{offset-POS}. \textit{TL} presents transliterations of the words in \textit{arb}. The \emph{numWordnets} is 4 and the  \emph{numCandidates} is 7. The rank of each candidate is shown in the last column of Figure \ref{fig:exampleRanking}.} 
\label{fig:exampleRanking}
\end{figure}

For instance, in our experiments we use 4 intermediate Wordnets, viz., PWN, FWN, JWN and WOLF Wordnet (WWN) \cite{Sagot2008}. The words in the \emph{offset-POS} "00006802-v" obtained from all 4 Wordnets, their translations to \emph{arb}, the occurrence count and the rank of each translation are presented in the second, the fourth and the fifth columns, respectively, of Figure \ref{fig:exampleRanking}.

\vspace{-2mm}

\subsection{Selecting candidates based on ranks}

We separate candidates based on three cases as below.

\begin{figure}[!h]
\centering \includegraphics[width=0.50\textwidth]{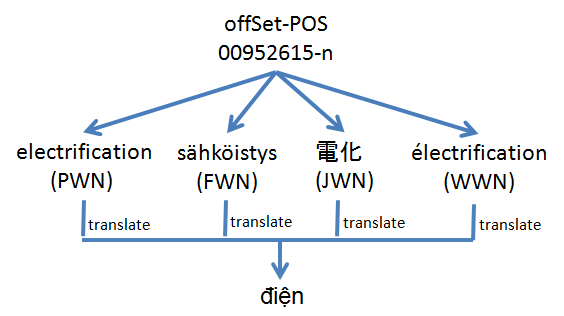}\vspace{-0.05 in}
\caption{Example of Case 1: Using the IW approach with four intermediate Wordnets, PWN, FWN, JWN and WWN. All words belonging to the \emph{offSet-POS} "00952615-n" in all 4 Wordnets are translated to the same word "{\fontencoding{T5}\selectfont \dj i\d\ecircumflex n}" in \emph{vie}. The word "{\fontencoding{T5}\selectfont \dj i\d\ecircumflex n}" is accepted as the correct word belonging to the \emph{offSet-POS} "00952615-n" in the Vietnamese Wordnet we create.}
\label{fig:case1}
\end{figure}

\textbf{Case 1:} A candidate \emph{w} has the highest chance to become a correct word belonging to a specific synset in the target language if its rank is 1.0. This means that all intermediate Wordnets contain the synset having a specific \emph{offset-POS} and all words belonging to these synsets are translated to the same word \emph{w}. The more the number of intermediate Wordnets used, the higher the chance the candidate with the rank of 1.0 has to become the correct translation. Therefore, we accept all translations that satisfy this criterion. An example of this scenario is presented in Figure \ref{fig:case1}.
 \vspace{-4mm}
 \vspace{-4mm}

\begin{figure}[!h]
\centering \includegraphics[width=0.50\textwidth]{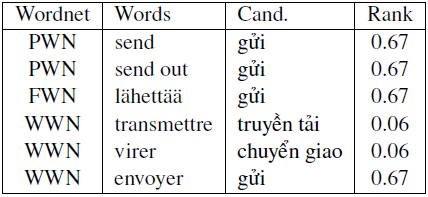}\vspace{-0.05 in}
\caption{Example of Case 2: Using the IW approach with three intermediate Wordnets, PWN, FWN and WWN. For the \emph{offSet-POS} "01437254-v", there is no candidate with the rank of 1.0. The highest rank of the candidates in "vie" is 0.67 which is the word {\fontencoding{T5}\selectfont g\h\uhorn i}. We accept "{\fontencoding{T5}\selectfont g\h\uhorn i}" as the correct word in the \emph{offSet-POS} "01437254-v" in the Vietnamese Wordnet we create.}
\label{fig:case2}
\end{figure}

\textbf{Case 2:} If an \emph{offSet-POS} does not have candidates having the rank of 1.0, we accept the candidates having the greatest rank. Figure \ref{fig:case2} shows the example of the second scenario.

\textbf{Case 3:}  If all candidates of an \emph{offSet-POS} has the same rank which is also the greatest rank, we skip these candidates. Table \ref{tab:Case3} gives an example of the last scenario.
\begin{table}[!h]
 \centering
\begin{tabular}{|c|l|l|c|}
\hline    
Wordnet& Words & Cand. & Rank\\
\hline    
PWN &act&{\fontencoding{T5}\selectfont h\`anh \dj\d\ocircumflex ng}&0.33\\
PWN &behave&{\fontencoding{T5}\selectfont ho\d{a}t  \dj\d\ocircumflex ng}&0.33\\
FWN &do&{\fontencoding{T5}\selectfont l\`am}&0.33\\
 \hline
\end{tabular}
\caption{Example of Case 3: Using the DR approach. For the \emph{offSet-POS} "00010435-v", there is no candidate with the rank of 1.0. The highest rank of the candidates in \textit{vie} is 0.33. All of 3 candidates have the rank as same as the highest rank. Therefore, we do not accept any candidate as the correct word in the \emph{offSet-POS} "00010435-v" in the Vietnamese Wordnet we create.}
\label{tab:Case3}
\end{table}

\vspace{-4mm}
\section{Experiments} 
\subsection{Publicly available Wordnets}
The PWN is the oldest and the biggest available Wordnet. It is also free. Wordnets in many languages are being constructed and developed\footnote{http://www.globalWordnet.org/gwa/Wordnet\_table.html}. However, only a few of these Wordnets are of high quality and free for downloading. The EuroWordnet \cite{Vossen1998} is a multilingual database with Wordnets in European languages (e.g., Dutch, Italian and Spanish). The AsianWordnet\footnote{http://www.asianWordnet.org/progress} provides a platform for building and sharing Wordnets for Asian languages (e.g., Mongolian, Thai and Vietnamese). Unfortunately, the progress in building most of these Wordnets is slow and they are far from being finished.

In our current experiments as mentioned earlier, we use the PWN and other Wordnets linked to the PWN 3.0 provided by the Open Multilingual Wordnet\footnote{http://compling.hss.ntu.edu.sg/omw/} project \cite{Bond2013}: WWN, FWN and JWN. Table \ref{tab:WordnetsInformation} provides some details of the Wordnets used.

\begin{table}[h]
 \centering
\begin{tabular}{|l|c|c|}
\hline    
\textbf{Wordnet }& \textbf{Synsets} & \textbf{Core}	\\
\hline    
JWN & 57,179 & 95\%\\ 
FWN &116,763 & 100\%\\
PWN& 117,659 & 100\%\\
WWN & 59,091 &92\%\\
 
\hline   
\end{tabular}
\caption{The number of synsets in the Wordnets linked to the PWN 3.0 are obtained from the Open Multilingual Wordnet, along with the percentage of synsets covered from the semi-automatically compiled list of 5,000 "core" word senses in PWN. Note that synsets which are not linked to the PWN are not taken into account.} 
\label{tab:WordnetsInformation}
\end{table}

For languages not supported by MT, we use three additional bilingual dictionaries: two dictionaries \emph{Dict(eng,ajz}) and \emph{Dict(eng,dis)} provided by Xobdo\footnote{http://www.xobdo.org/}; one \emph{Dict(eng,asm)} created by integrating two dictionaries \emph{Dict(eng,asm)} provided by Xobdo and Panlex\footnote{http://panlex.org/}. The dictionaries are of varying qualities and sizes. The total number of entries in \emph{Dict(eng,ajz)}, \emph{Dict(eng,asm)} and \emph{Dict(eng,dis)} are 4682,  76634 and 6628, respectively. 
 
\subsection{Experimental results and discussion}
As previously mentioned, our primary goal is to build high quality synsets for Wordnets in languages with low amount of resources: \emph{ajz, asm, arb, dis} and \emph{vie}. The number of Wordnet synsets we create for \textit{arb} and \textit{vie} using the DR approach and the coverage percentage compared to the PWN synsets are 4813 (4.10\%) and 2983 (2.54\%), respectively.
The number of synsets for each Wordnet we create using the IW approach with different numbers of intermediate Wordnets and the coverage percentage compared to the PWN synsets are presented in Table \ref{tab:NumSynsetIW}. 
\begin{table}[!h]
 \centering
\begin{tabular}{|c|c|c|r|c|}
\hline    
\textbf{App.}&\textbf{Lang.}& \textbf{WNs}&\textbf{Synsets}&\textbf{\% coverage} \\
\hline    
IW &arb&2& 48,245&41.00\%\\ 
IW &vie&2&42,938&36.49\%\\
IW &arb&3&61,354&52.15\%\\ 
IW &vie&3&57,439&48.82\%\\
IW &arb&4&75,234&63.94\%\\ 
IW &vie&4&72,010&61.20\%\\ 
\hline   
\end{tabular}
\caption{The number of Wordnet synsets we create using the IW approach. \textit{WNs} is the number of intermediate Wordnets used: 2: PWN and FWN, 3: PWN, FWN and JWN and 4: PWN, FWN, JWN and WWN.} 
\label{tab:NumSynsetIW}
\end{table}
 
For the IWND approach, we use all 4 Wordnets as intermediate resources. The number of Wordnet synsets we create using the IWND approach are presented in Table \ref{tab:NumSynsetIWND}. We only construct Wordnet synsets for \emph{ajz}, \emph{asm} and \emph{dis} using the IWND approach because these languages are not supported by MT.
\vspace{-0.08 in}
\begin{table}[!h]
 \centering
\begin{tabular}{|c|c|r|c|}
\hline    
\textbf{App.}&\textbf{Lang.} &\textbf{Synsets}&\textbf{\% coverage} \\
\hline    
IWND &ajz&21,882&18.60\%\\ 
IWND &arb&70,536&59.95\%\\ 
IWND &asm&43,479&36.95\%\\ 
IWND &dis&24,131&20.51\%\\ 
IWND &vie&42,592&36.20\%\\ 
\hline   
\end{tabular}
\caption{The number of Wordnets synsets we create using the IWND approach.}
\label{tab:NumSynsetIWND}
\end{table}
 
Finally, we combine all of the Wordnet synsets we create using different approaches to generate the final Wordnet synsets. Table \ref{tab:FinalSynset} presents the final number of Wordnet synsets we create and their coverage percentage.
\vspace{-0.08 in}
\begin{table}[!h]
 \centering
\begin{tabular}{|c|r|c|}
\hline    
\textbf{Lang.} &\textbf{Synsets} & \textbf{\% coverage}\\
\hline    
ajz&21,882&18.60\%\\ 
arb&76,322&64.87\%\\
asm&43,479&36.95\%\\
dis&24,131&20.51\%\\ 
vie&98,210&83.47\%\\
\hline   
\end{tabular}
\caption{The number and the average score of Wordnets synsets we create.}
\label{tab:FinalSynset}
\end{table}
\vspace{-0.1 in}

Evaluations were performed by volunteers who use the language of the Wordnet as mother tongue. To achieve reliable judgment, we use the same set of 500 \emph{offSet-POS}s, randomly chosen from the synsets we create. Each volunteer was requested to evaluate using a 5-point scale -- 5: excellent, 4: good, 3: average, 2: fair and 1: bad. The average score of Wordnet synsets for \textit{arb, asm} and \textit{vie} are 3.82, 3.78 and 3.75, respectively. We notice that the Wordnet synsets generated using the IW approach with all 4 intermediate Wordnets have the highest average score: 4.16/5.00 for \emph{arb} and 4.26/5.00 for \emph{vie}. We are in the process of finding volunteers to evaluate the Wordnet synsets for \textit{ajz} and \textit{dis}.
  
It is difficult to compare Wordnets because the languages involved in different papers are different, the number and quality of input resources vary and the evaluation methods are not standard. However, for the sake of completeness, we make an attempt at comparing our results with published papers. Although our score is not in terms of percentage, we obtain the average score of 3.78/5.00 (or informally and possibly incorrectly, 75.60\% precision) which we believe it is better than 55.30\% obtained by \cite{Bond2008} and 43.20\% obtained by \cite{Charoenporn2008}. In addition, the average coverage percentage of all Wordnet synsets we create is 44.85\% which is better than 12\% in \cite{Charoenporn2008} and 33276 synsets ($\simeq$ 28.28\%) in \cite{Saveski2010} .

The previous studies need more than one dictionary to translate between a target language and intermediate-helper languages. For example, to create the JWN, \cite{Bond2008} needs the Japanese-Multilingual dictionary, Japanese-English lexicon and Japanese-English life science dictionary. For \textit{asm}, there are a number of Dict(eng,asm); to the best of our knowledge only two online dictionaries, both between \textit{eng} and \textit{asm}, are available. The IWND approach requires only one input dictionary between a pair of languages. This is a strength of our method.

\section{Conclusion and future work}
We present approaches to create Wordnet synsets for languages using available Wordnets, a public MT and a single bilingual dictionary. We create Wordnet synsets with good accuracy and high coverage for languages with low resources (\textit{arb} and \textit{vie}), resource-poor (\textit{asm}) and endangered (\textit{ajz} and \textit{dis}). We believe that our work has the potential to construct full Wordnets for languages which do not have many existing resources. We are in the process of creating a Website where all Wordnet synsets we create will be available, along with a user friendly interface to give feedback on individual entries. We will solicit feedback from communities that use these languages as mother-tongue. Our goal is to use this feedback to improve the quality of the Wordnet synsets. Some of Wordnet synsets we created can be downloaded from http://cs.uccs.edu/$\thicksim$linclab/projects.html.

\end{document}